\documentclass[10pt,twocolumn,letterpaper]{article}

\usepackage{iccv}
\usepackage{times}
\usepackage{epsfig}
\usepackage{graphicx}
\usepackage{amsmath}
\usepackage{amssymb}

\usepackage[pagebackref=true,breaklinks=true,letterpaper=true,colorlinks,bookmarks=false]{hyperref}

\usepackage{xcolor}
\usepackage{placeins}
\usepackage{amsmath}
\usepackage{enumitem}

\iccvfinalcopy 

\def\iccvPaperID{SG2RL-16} 

\ificcvfinal\fi

\begin{document}

\title{Haystack: A Panoptic Scene Graph Dataset to Evaluate Rare Predicate Classes}

\author{Julian Lorenz\hskip 1.5em Florian Barthel\hskip 1.5em Daniel Kienzle\hskip 1.5em Rainer Lienhart\\
University of Augsburg\\
Augsburg, Germany\\
{\tt\small \{julian.lorenz,florian.barthel,daniel.kienzle,rainer.lienhart\}@uni-a.de}
}

\maketitle
\ificcvfinal\thispagestyle{empty}\fi

\begin{abstract}
Current scene graph datasets suffer from strong long-tail distributions of their predicate classes. Due to a very low number of some predicate classes in the test sets, no reliable metrics can be retrieved for the rarest classes. We construct a new panoptic scene graph dataset and a set of metrics that are designed as a benchmark for the predictive performance especially on rare predicate classes. To construct the new dataset, we propose a model-assisted annotation pipeline that efficiently finds rare predicate classes that are hidden in a large set of images like needles in a haystack.

Contrary to prior scene graph datasets, Haystack contains explicit negative annotations, \ie annotations that a given relation does \textbf{not} have a certain predicate class. Negative annotations are helpful especially in the field of scene graph generation and open up a whole new set of possibilities to improve current scene graph generation models.

Haystack is 100\% compatible with existing panoptic scene graph datasets and can easily be integrated with existing evaluation pipelines. Our dataset and code can be found here: \url{https://lorjul.github.io/haystack/}. It includes annotation files and simple to use scripts and utilities, to help with integrating our dataset in existing work.
\end{abstract}

\section{Introduction}

In scene graph generation, models are trained to detect and classify interactions between objects in an image. These interactions are called relations and are composed of three components: subject, predicate, and object. Existing methods have improved over the last years but are still struggling with the long-tail distribution of the predicate classes in scene graph datasets \cite{survey2023,survey2022} and therefore perform worse on rare predicates.

Much research is conducted to find methods that can tackle the long-tail problem of scene graph datasets. Although these methods can reduce the performance gap between head and tail classes, they are still limited by the lack of available relations with tail predicates in existing datasets. For example, due to very small test sets, existing methods cannot be reliably evaluated on rare predicates. Additionally, commonly used metrics from the Recall@k family can only provide insights on an image-level, without paying too much attention on a per relation basis.

\begin{figure}
    \centering
    \includegraphics[width=\linewidth]{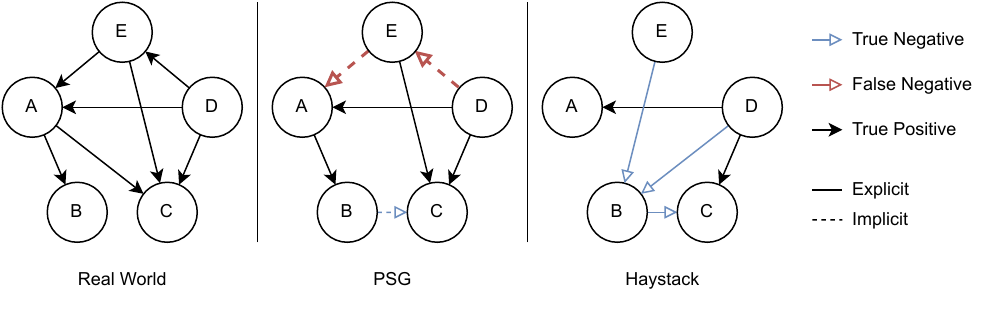}
    \caption{Schematic comparison of the different annotation structures for our Haystack dataset and the PSG dataset. Our dataset prefers more annotations for rare predicates over full annotations of an image. Additionally, our dataset contains explicit negative annotations which must be implicitly derived for PSG.}
    \label{fig:intro-fig}
\end{figure}

We define a new set of metrics that can evaluate relations individually and provide substantial new information about existing methods. Our metrics can grade the model's understanding of a specific predicate as well as influences between predicates before they are ranked for the final inference output.

However, our metrics require reliable annotations, including negative annotations. Negative annotations show which predicates are \emph{not} part of a specific relation. These annotations are not explicitly given for current scene graph datasets, preventing in-depth analysis on current test sets.

To address this issue, we construct a new panoptic scene graph dataset that includes explicit negative annotations for rare predicate classes. Because existing test sets are rather lacking for rare predicate classes, we decide to create a new test dataset from scratch. Contrary to most prior scene graph datasets, our dataset is not a subset of Visual Genome \cite{visualgenome} but SA-1B \cite{sam}. Therefore, our dataset can simply be used in addition to existing training and evaluation pipelines without having to deal with overlapping datasets.

We use an efficient annotation pipeline that is designed to get many annotations for rare predicate classes as fast as possible. Therefore, we first identify two main problems why existing scene graph datasets struggle with tail classes:

\begin{enumerate}[topsep=0pt, itemsep=0pt]
    \item Annotators are tasked to annotate one image after the other without prioritizing images with potential rare predicates.
    \item Annotators look at the image as a whole and have a set of predicates to choose from and tend to select more basic predicates instead the more informative but rare predicates.
\end{enumerate}

We address these problems using a model-assisted annotation pipeline that searches through a large amount of 11 million images from the SA-1B \cite{sam} dataset and retrieves promising candidates for manual annotation.

Our dataset is directly compatible with existing panoptic scene graph datasets and can be easily integrated with existing evaluation pipelines.

Our main contributions are:

\begin{enumerate}[topsep=0pt, itemsep=0pt]
    \item An active learning inspired annotation pipeline that can be used to efficiently build scene graph datasets with a focus on rare predicate classes. We use model-assisted proposals to find rare predicate classes in a large set of unlabeled images.
    \item With our pipeline, we build the Haystack scene graph dataset that contains about 25,000 relations with rare predicate classes for more than 11,300 images. It includes negative annotations and can be used for better model evaluation on rare predicate classes.
    \item A set of metrics that provide more in-depth insights into results on rare predicates and which are used to compare existing approaches.
\end{enumerate}

\section{Related Works}

\subsection{Scene Graph Datasets}
\label{sec:datasets}

One of the first large scene graph datasets used for scene graph generation is Visual Genome \cite{visualgenome}. It contains more than 100,000 images but has some ill-suited properties, \eg 33,877 different object classes and 40,480 different predicate classes. These classes are mostly raw labels by annotators with only very slight data post processing. To improve this, Xu \etal took Visual Genome and constructed the commonly used VG-150 \cite{imp} variant, keeping only the most frequent 50 predicate classes and 150 object classes. Although this variant drastically reduced the number of different predicate classes to the most relevant ones, many predicates are still redundant.

Yang \etal identified these issues and created the PSG \cite{psg} dataset. It is based on the intersection of images from Visual Genome and COCO and contains 48749 images with panoptic segmentation masks and a total of 56 predicate classes. The authors tackled the issues of prior scene graph datasets and decided to use a completely new set of predicate classes. They focused on a less redundant predicate vocabulary that can still be used to concisely represent the given scene as thorough as possible. Annotators were then given the fixed set of predicates and encouraged to use more informative predicates whenever applicable. Additionally, the authors made sure that not only salient regions of an image were covered with relation annotations.

Still, predicate classes in the PSG dataset follow a long-tail distribution like prior datasets. However, compared to Visual Genome, PSG contains more reliable annotations and we decide to choose this dataset as the training set for our new test set.

\begin{figure}
    \centering
    \includegraphics[width=\linewidth]{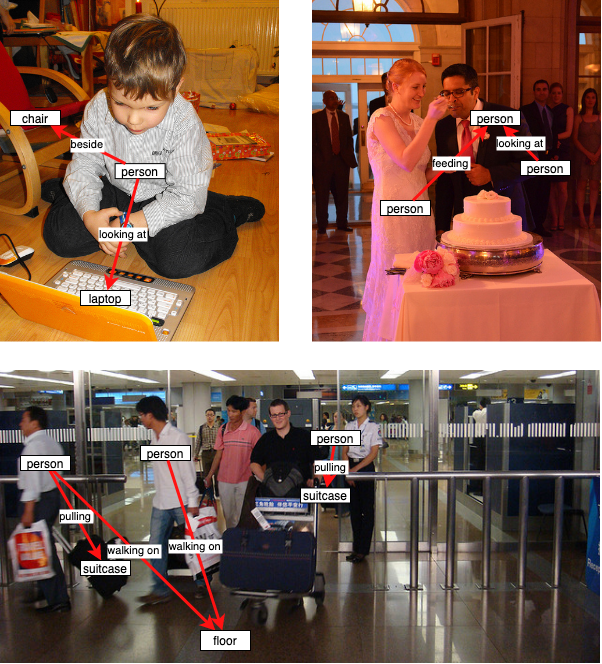}
    \caption{Example images with annotations that were missed in the PSG ground truth.}
    \label{fig:bad-psg-images}
\end{figure}

Annotating scene graphs extensively is very difficult. Although annotations for PSG are much more complete compared to Visual Genome, there are still a lot of images with many missing annotations. See figure \ref{fig:bad-psg-images} for example images where annotations were missed.

Our annotation pipeline reuses the predicate classes from the PSG dataset but adds a whole new set of images, containing mostly rare predicate classes. Generating exhaustive scene graph datasets is a near impossible task and we decide to go a different route with the Haystack dataset. Instead of focusing only on positive annotations, we include negative annotations as well (figure \ref{fig:intro-fig}). We will argue in section \ref{sec:evaluation} why this dataset structure is superior when evaluating individual rare predicates classes.

\subsection{Scene Graph Generation with Long-Tail Data}

Both Visual Genome and derivatives like PSG suffer from a long-tail distribution of the predicate classes. In the case of PSG, the 3 most frequent predicate classes "on", "beside", and "over" make up 52\% of all available predicate labels in the dataset. More than a quarter of all predicate classes have less than 100 annotations in the dataset.

This is a known problem in the field of scene graph generation and has been approached for example using resampling \cite{devil_in_tails}, reweighting \cite{pcpl}, or predicate grouping \cite{dong_stacked}.
Zhang \etal proposed a method to automatically relabel existing datasets during training and convert less informative annotations to more informative ones on the fly. Zhou \etal built on this work and developed a model that works with panoptic scene graph datasets \cite{HiLo}.

However, all of these methods have to evaluate on the same lacking test sets with very few samples for rare predicate classes. With our dataset, they can be evaluated on a more reliable test set.

\subsection{Metrics for Scene Graph Generation}

In scene graph datasets, ground truth annotations are incomplete, making it difficult to apply arbitrary metrics from other fields in machine learning. To use standard metrics like accuracy, positive labels and negative labels are required, too. However, scene graph datasets only contain positive annotations for the underlying relations. This is usually not a problem for classification tasks because normally, there exists exactly one label per data sample. For scene graph datasets, the situation is different. Here, relations can have zero, one, or even multiple predicate classes assigned. Consequently, if a predicate class is missing from a relation in the ground truth, it doesn't necessarily mean that the predicate is not suited for that relation. Quite to the contrary, many images from current scene graph datasets contain images with many missed annotations. Therefore, the lack of a predicate class can only serve as a guess for a negative annotation.

To cope with this problem, most work on scene graph generation uses Recall@k \cite{lu2016visual} or a variant of it. Recall@k is calculated at an image level. Starting from a model output tensor that contains one row per possible relation and one column for each available predicate class, the rows are ranked by their most confident predicate score. Next, given the set of ground truth relations, we can check how many relations are covered by the top k ranked predicates from the model output tensor and calculate a ratio for the image. Recall@k is the average over all these ratios. The Recall@k metric gives insight into how good the model is at filtering the most relevant relations on an image. However, its main disadvantage is that it favors frequent predicate classes over rare ones. On PSG, a hypothetical model that would only get all relations with the predicates "on", "beside", or "over" right, would already achieve a Recall@k of 0.42.

A metric that is better suited to analyze the performance with the long-tail distribution in mind is the mean Recall@k, a variant of Recall@k that first calculates individual scores for every available predicate class and averages the values afterwards. This way, every predicate class has the same influence towards the final metric score.
Another variant is the "No Graph Constraint Recall@k" \cite{pixels2graph,motifs} that allows multiple predicates per relation for the ranking.

Metrics from the Recall@k family provide insights into how good the tested model is at ranking relevant relations on an image.

\section{Methods}

\begin{figure*}
    \centering
    \includegraphics[width=\textwidth]{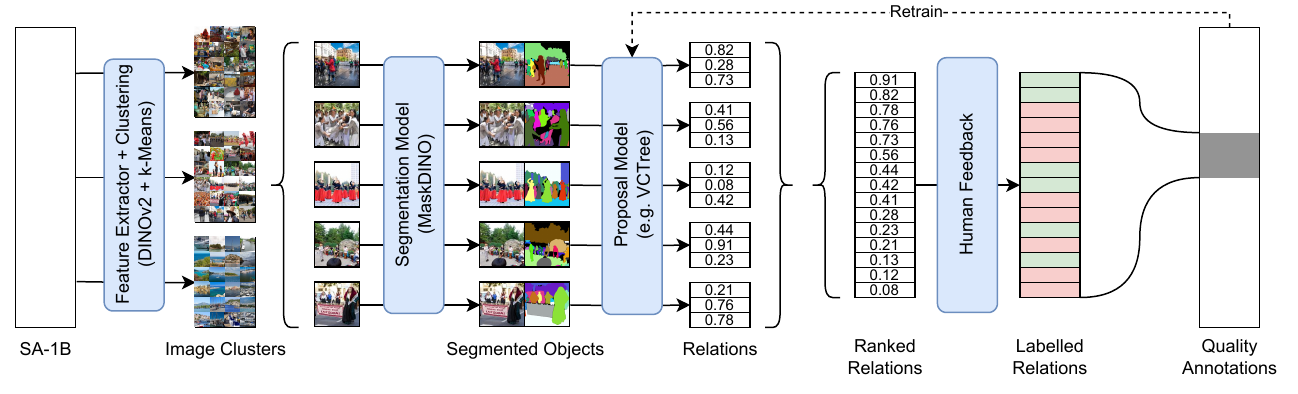}
    \caption{Overview of our annotation pipeline for Haystack. The pipeline is designed to find rare predicates in a very large set of images (the SA-1B dataset). We first cluster the images, to increase the diversity and then calculate segmentation masks for each image. These segmentation masks are compatible with the PSG dataset. Next, we apply a scene graph model on the set of images and let it predict scores for all possible relations on the images. The scores are ranked and annotated by hand. In contrast to existing scene graph datasets, this allows us to publish negative annotations as well, which can be used later for an improved training.}
    \label{fig:pipeline}
\end{figure*}

Traditionally, scene graph datasets are annotated on a per-image basis \cite{visualgenome,psg}. The annotator is tasked to annotate as many relations between objects as possible on a given image. To ensure a good quality of the annotations, annotators are encouraged to use more informative predicate classes whenever possible. However, this only works to a certain degree and the annotators must have a good overview over all available predicate classes to choose correctly. Therefore, we must change two fundamental steps in the annotation process to shift the focus to the tail classes:

First, images must be sorted by the estimated chance to find rare predicates. We use a model-assisted approach for this task.

Second, annotators should not have to keep the whole set of available predicates in mind. If given the choice, annotators tend to use more broader predicate classes like "on" or "beside" \cite{IETrans}. Therefore, we essentially reduce the annotation task to a binary one and use the proposal model to only show relations that are expected to have a given predicate class. In this case, the annotator only has to know about one predicate and is less likely to make any errors.

\subsection{Annotation Pipeline}
\label{sec:pipeline}

An overview of our annotation pipeline can be seen in figure \ref{fig:pipeline}. From a large image database, we first extract objects together with their segmentation masks using MaskDINO \cite{maskdino}. Next, we use these masks as ground truth data for inference with a pretrained scene graph generation model. We use this model as a proposal algorithm to select relations that are likely to contain rare predicate classes. Starting from there, annotators are tasked to verify the various predicates and label the proposed relations as either correct or incorrect. Contrary to prior scene graph datasets, we publish negative annotations, too, which opens the doors to a whole new set of training and evaluation techniques. To increase the diversity of our dataset, we cluster all available images in distinct groups and sample from them uniformly.

\textbf{Source images} Instead of extending existing scene graph datasets, we decide to start from scratch and introduce a completely new set of images for our scene graph dataset. To increase the chance of finding rare predicates, we process all images from the SA-1B \cite{sam} dataset. SA-1B contains more than 11 million high quality images from different domains. We iterate over all available images and filter them depending on a proposal algorithm that we will explain later.

\begin{figure}
    \centering
    \includegraphics[width=0.9\linewidth]{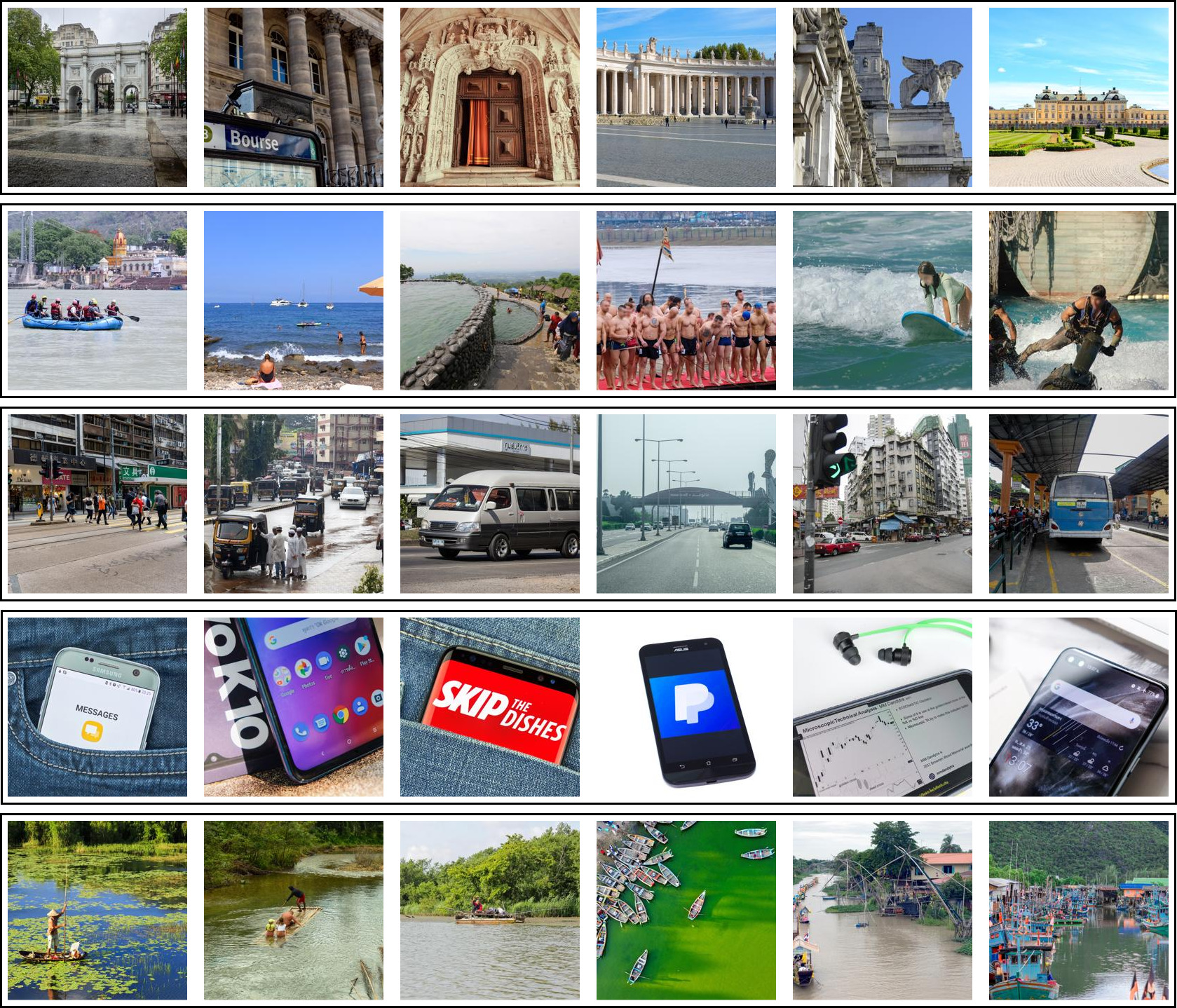}
    \caption{Example images of the first 5 clusters. Each row is a separate cluster. The clusters were calculated using k-Means with features from DINOv2 \cite{dinov2}. For some clusters, the contained images were not suitable for our task like the fourth cluster above.}
    \label{fig:cluster-examples}
\end{figure}

\textbf{Increase diversity} Because we use a trained neural network to propose new relations, there will be a tendency to a certain group of images. This would reduce the diversity of our dataset. However, generating a diverse dataset is important to improve the robustness of trained models. Thus, we first cluster the images from SA-1B based on features from DINOv2 \cite{dinov2}. DINOv2 is trained without supervision on a large set of images and produces features that can be used for further processing even without requiring a retraining of the backbone. We compute features with the ViT-L checkpoint. Next, we apply k-Means and put the images into 50 disjoint clusters. The number of clusters was empirically selected by iteratively changing the number of clusters on a smaller set of images. 50 clusters is a convenient tradeoff that produces diverse clusters that still contain varying images inside them. Some example images for the first 5 clusters are shown in figure \ref{fig:cluster-examples}. Not all images from SA-1B are suitable candidates for our scene graph dataset, \eg logos, or portraits. We manually inspect example images for each cluster and decide whether to exclude all images from a cluster.

We use the remaining clusters as pools for our proposal algorithm. To propose new relations for annotation, we first sample uniformly from the set of clusters and for each selected cluster, we apply the model-guided proposal algorithm to rank the most promising candidates for manual annotation. With the combination of both clusters and network guided proposals, we make sure that we generate relevant relation candidates which are based on diverse images.

\textbf{Segmentation masks} The images from SA-1B are not compatible with PSG because they are lacking the required panoptic segmentation masks. There are segmentation masks available, but they were extracted with SAM \cite{sam} and don't resemble the 133 thing and stuff classes from PSG. Annotating the missing segmentation masks by hand would be very inefficient and error prone. Therefore, we use MaskDINO \cite{maskdino}, trained on the object classes of PSG and collect predictions for the full SA-1B dataset. MaskDINO is a foundation model, capable to do object detection and segmentation. It achieves state of the art results on COCO instance segmentation and indeed, the vast majority of the returned segmentation masks for our task is almost pixel-perfect and suitable for further processing. Masks that are not good enough will be filtered later in our annotation pipeline.

\begin{figure}
    \centering
    \includegraphics[width=0.95\linewidth]{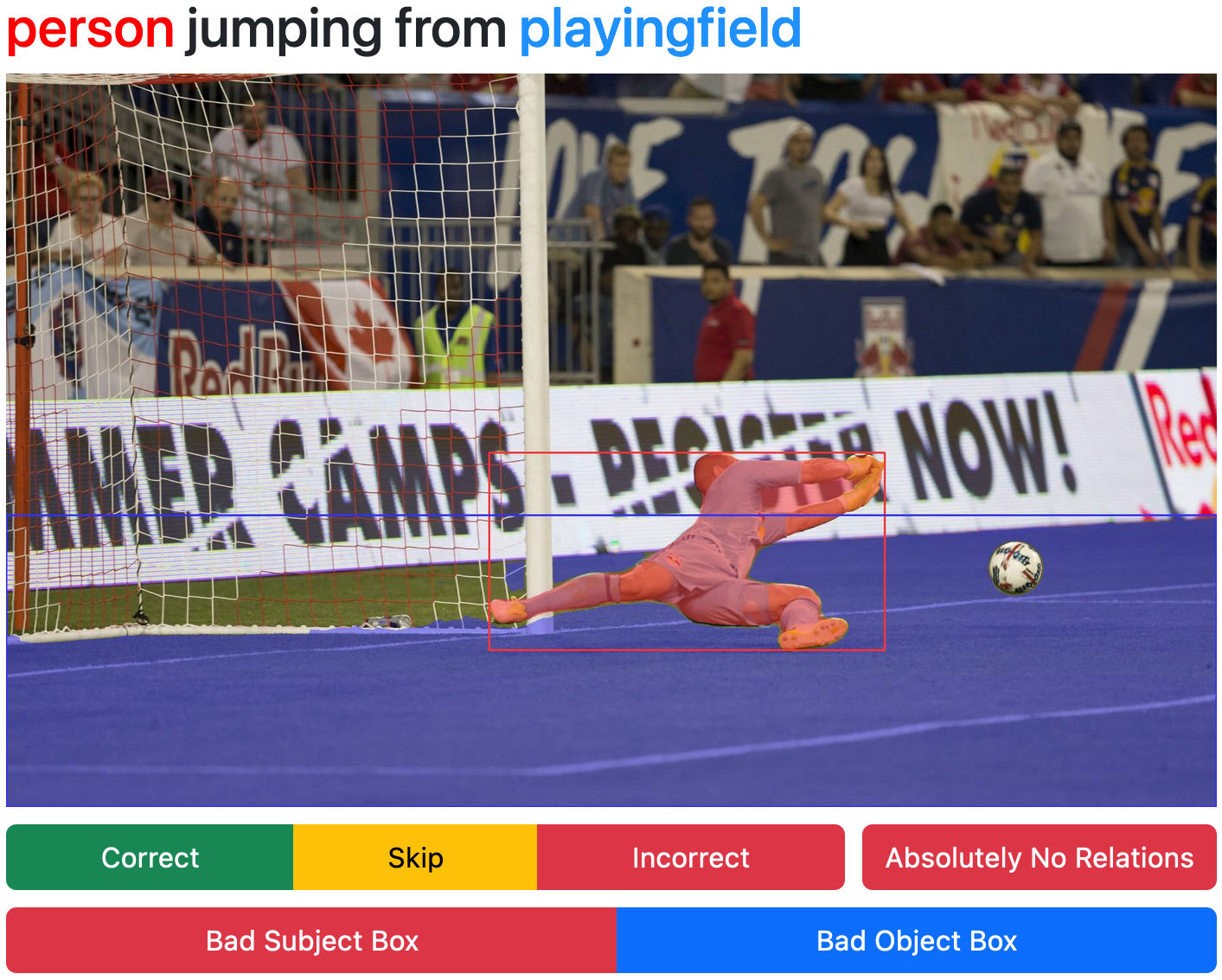}
    \caption{A screenshot of our annotation interface. The annotator is given a fixed predicate to label, in this case "jumping from". On every image, one subject and one object are highlighted in different colors. The annotator can choose to classify the proposal as correct/incorrect. Additionally, if the segmentation mask would have errors, there are two buttons to exclude the respective subject or object from further proposals. In this example, the annotator would click on "correct".}
    \label{fig:interface}
\end{figure}

\textbf{Predicate renaming} We observe that annotators that are not familiar with the PSG dataset have difficulties to apply the selected predicate class definitions. This is due to misunderstandings of the predicate classes when translated to other languages. To ensure that our test set is 100\% aligned with the definitions from PSG, we decide to rename the existing predicate classes for our annotators.

For every predicate class, we select a set of images that contain at least one relation with the given predicate. Next, without showing the actual list of predicate classes from PSG, we let the annotators decide on a predicate label for the given set of relations. Annotators are free to describe the relation in their own words how they feel it would fit best. Afterwards we check if the proposed new predicate name does indeed describe the predicate class.

During this process, annotators for example renamed the predicate class "playing" to "engaged in activity using". This new label makes the difference between the PSG predicates "playing" and "playing with" much more evident for our annotators. For the final dataset, we convert the renamed labels back to the original ones.

\textbf{Annotation interface} To add relation labels for the given images and their PSG-compatible segmentation masks, annotators could just label one image after the other and select all visible relations until enough data is available. However, there are two disadvantages to this approach: First, it is very inefficient to provide extensive annotations for each image. Our pre-processed images contain on average 16 objects per image, which would result in about 240 possible relations per image. But second and more importantly, annotators would not focus on rare predicate classes if they annotated the image as a whole.

We actively prevent this phenomenon by fixing the predicate and showing potential relation candidates one after the other. The annotator can label the relation with one of three choices (top row in figure \ref{fig:interface}):

\begin{enumerate}[topsep=0pt, itemsep=0pt]
    \item Positive annotation: the fixed predicate does in fact fit the proposed relation.
    \item Negative annotation: the fixed predicate does not fit the proposed relation, but another predicate would fit. To speed up the annotation process, the annotator does not label the correct positive annotation.
    \item No relation: there is no predicate that would fit the proposed relation. This is a shortcut to applying option 2 for all predicates.
\end{enumerate}

Regardless of the outcome, we store both positive and negative annotations for later use.
Additionally, annotators can decide to skip a proposed image if they are not sure about the annotation.

\textbf{Model-assisted proposals} We use a scene graph generation model to propose probable relation candidates to the annotators. Because our pipeline does not depend on a specific choice of model, we choose the top-performing model VCTree \cite{vctree} from the PSG paper\cite{psg}. We train it on the original PSG dataset and then calculate all possible relations between all available objects in all available images from the selected cluster and calculate a score for each predicate class. The score is normalized with the softmax function to prevent the model from focusing too much on certain images. Given these scores, we can rank the processed relation candidates.



It is worth noting, that most scene graph models contain a dedicated output for "no relation". In order to focus on proposals that are likely to show a relation, we rank our proposals by dividing the predicted score for a predicate by the predicted score for "no relation". Hence, relations with high "no relation" scores, are shown less frequently.

\textbf{Additional annotation options} Although MaskDINO provides impressive segmentation masks for our dataset, the masks are not always correct. In this case, annotators have the choice to mark objects as faulty and exclude them from further processing (bottom row in figure \ref{fig:interface}). We will not use these objects for training or evaluation.

\textbf{Filter nonsense} To improve relation selection, we filter subject-predicate-object triplets unlikely to be viable for our new dataset. For instance, annotations like "table-drinking-water" are eliminated. We use PSG statistics to count how often a subject appears with a predicate, regardless of the object. Subject-predicate combinations with one or zero samples in PSG are considered noise and excluded. The same applies to predicate-object combinations. Although this reduces dataset diversity, our proposal algorithm can still suggest never-before-seen subject-predicate-object triplets if either subject-predicate or predicate-object pairs exist in PSG. For example, PSG contains relations with "dog-eating" and "eating-banana" but not "dog-eating-banana". Note that out-of-set triplets are less likely due to using proposals from PSG pretrained models.

\textbf{Retrain the proposal model} Finally, we use our found annotations to retrain the proposal model during the annotation process. Once the new proposal model is trained, we use it as an improved proposal algorithm.

\subsection{Dataset Properties}

Our Haystack dataset is designed to contain as many rare predicates as possible, to provide a reliable test set for rare predicate classes. The dataset can be easily combined with existing scene graph datasets, such as PSG. To ensure compatibility, we reuse the predicate classes from PSG, but focus on the tail classes.

\begin{figure}
    \centering
    \includegraphics[width=\linewidth]{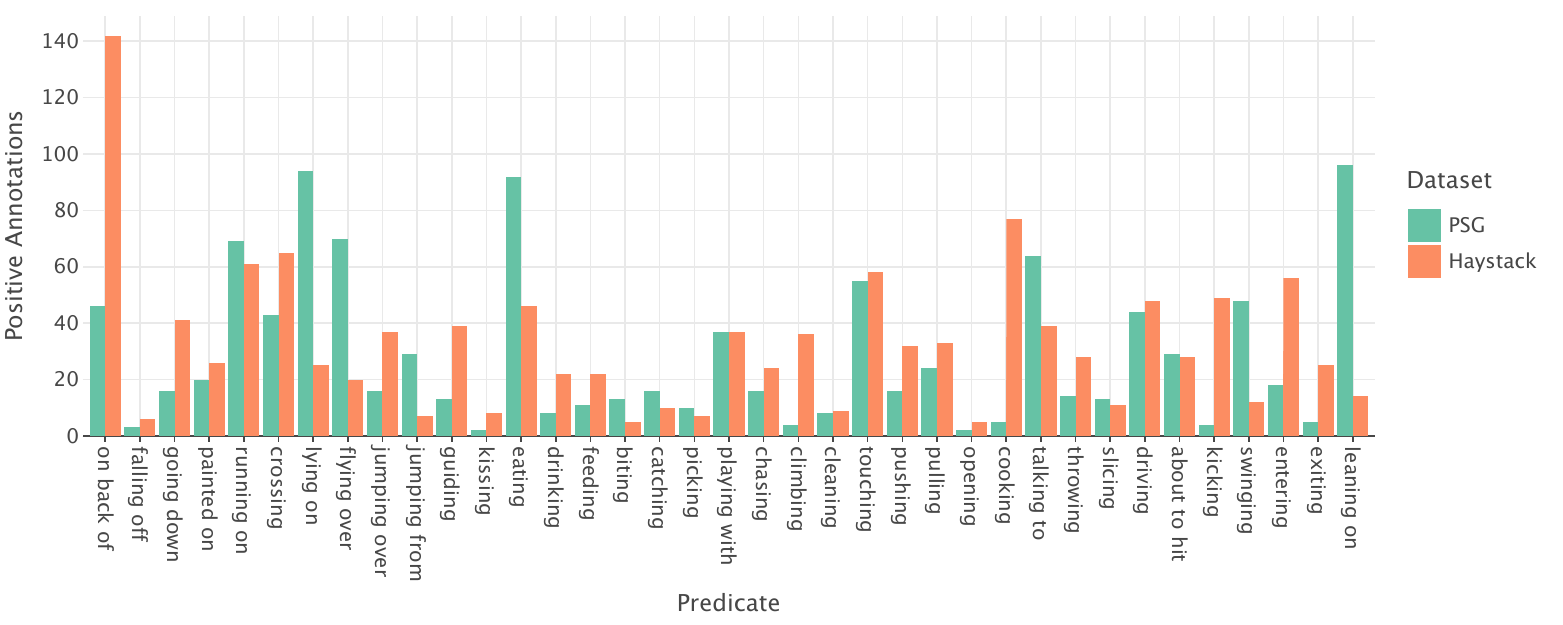}
    \caption{Distribution over all positive annotations in our dataset and the PSG test set. Our dataset contains more labels for almost all rare predicates when compared with the PSG test set.}
    \label{fig:data-distr}
\end{figure}

\begin{figure}
    \centering
    \includegraphics[width=\columnwidth]{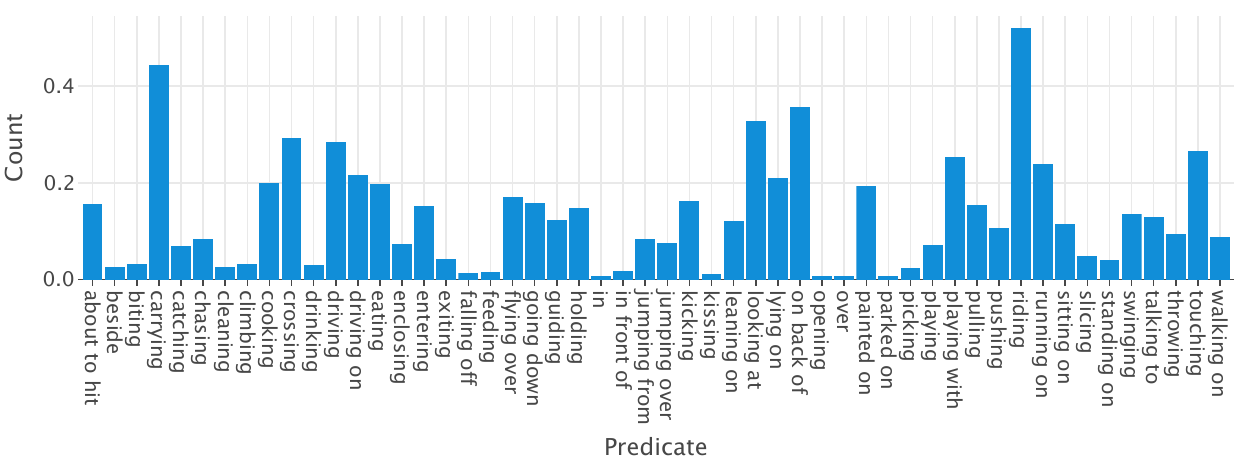}
    \caption{Percentage of how many annotations per predicate are positive in the dataset. Some predicates are easier to find because the used model proposals are better. For example, riding can be easily retrieved.}
    \label{fig:data-posratio}
\end{figure}

Haystack contains more than 25,000 relation annotations on a total of more than 11,300 images. Using our annotation pipeline, annotators were able to find 9\% positive annotations out of all proposed annotations for rare predicate classes in total (see figure \ref{fig:data-posratio} for per predicate ratios). Figure \ref{fig:data-distr} shows a list of all positive annotations in our dataset compared with the PSG test set.

\begin{figure}
    \centering
    \includegraphics[width=\columnwidth]{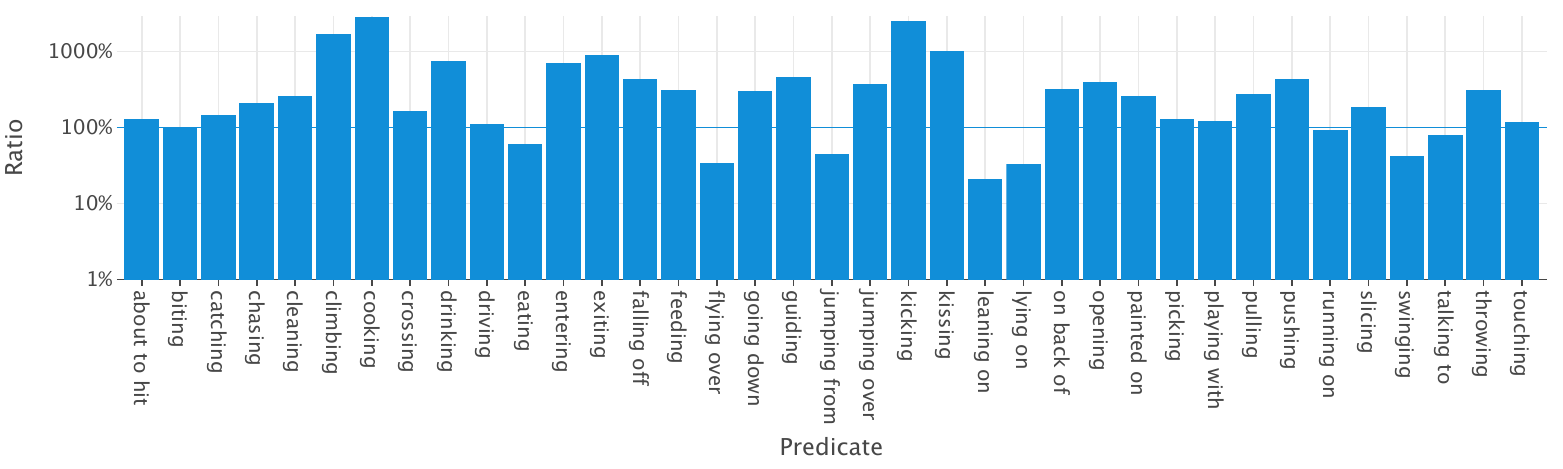}
    \caption{Relative dataset size of Haystack compared to the PSG test set (log scaled).}
    \label{fig:data-stats}
\end{figure}

Contrary to PSG, we annotate on a per-predicate basis instead of a per-image one. Consequently, the annotation density per image is lower compared to PSG. But at the same time, Haystack contains more rare predicate classes than PSG. For example, Haystack contains more than 10 times more relations with the predicate "cooking" or "climbing". See figure \ref{fig:data-stats} for the sample size increase on the rarest predicate classes.

For every image that contains at least one annotated relation, we provide the respective segmentation mask with the same resolution as SA-1B, that is, 1500 pixels for the shorter edge. Additionally, we provide an annotation file that uses the same file format as PSG and is 100\% compatible. Haystack can be effortlessly integrated into current PSG-based scene graph pipelines by appending our annotations to the existing JSON annotation files. We provide a small utility script in our repository that facilitates this step even further.

\subsection{Evaluation with the Haystack Dataset}
\label{sec:evaluation}

Previous work usually evaluates its results using the Recall@k (R@k) and Mean Recall@k (mR@k) metrics. R@k ranks all relation predictions for an image and returns the ratio of how many ground truth annotations are covered by the top k ranked predictions. The score is then averaged over all images. This design choice is required because only positive annotations are available in previous datasets. If a missing annotation between a subject and an object would imply a negative annotation, the model output would be compared with many false negative ground truth values. For the final R@k score, predicates are essentially competing with each other for the top ranked positions.

R@k scores two different aspects at the same time: the model's capability of recognizing predicates for different relations and ranking them by relevance for the final output. This makes sense for final evaluation but does not provide fine grained insights into a model's performance. Therefore, we define three metrics that evaluate the two aspects separately.

A fundamental requirement of the proposed metrics is the availability of negative annotations. For prior scene graph datasets, these could be derived from positive annotations or subject-object pairs that have no annotation. However, as mentioned in section \ref{sec:datasets}, relying on implicit negative annotations does not provide a reliable base for metrics. With the Haystack dataset, explicit negative annotations become available and our new metrics can be calculated without the risk of noisy ground truths due to implicit negative annotations.


Our metrics can be used to analyze different aspects of model performance. A usual inference task for scene graph generation is to process an input image and return a list of all visible relations in the image. Using R@k, we can calculate a score that represents how successful a model can achieve this task. However, R@k only looks at the bigger picture.

We design the \emph{Predicate ROC-AUC} (P-AUC) score as the ROC-AUC over individual predicate scores. More precisely, to calculate the P-AUC for a fixed predicate class $p$, we first collect all relations that have a positive or negative ground truth annotation for $p$. Next, we calculate the corresponding predictions for each relation and only look at the scores that relate to $p$. We now have a list of confidences and a list of labels and can calculate the ROC-AUC. The ROC-AUC has some beneficial properties for our task: It is invariant to scale and transformation and can, therefore, score any predicate regardless of the average confidence. This is important because many predicate classes like "carrying" or "pushing" are often predicted with very low scores compared to other predicates.
P-AUC describes the model's capability to decide whether a predicate class is applicable to a relation, regardless of the predictions for other predicates.

To understand how the predicate scores interfere with each other, we define two displacement metrics: \emph{Predicate Dominance Overestimation} (PDO) scores how much a predicate displaces other predicates, whereas \emph{Predicate Discrimination Disadvantage} (PDD) determines how much a predicate is displaced by other predicates. Both metrics are defined for a fixed predicate $p$.

Let $n$ be the number of possible predicate classes. Each ground truth annotation of a relation can be represented as a vector $l \in \{0,1,-1\}^n$ ($l_p = 0$ if there is no annotation for this relation, $l_p = 1$ if $p$ would be a correct predicate class for the relation, $l_p = -1$ if not). For every relation $l$, a model ranks the predicate classes by their confidence scores (low rank corresponds to high confidence): $r \in [0,n - 1]^n \subset \mathbb{N}^n$. Let $R_p$ be the set of annotated relations for predicate $p$: $R_p = \left\{ (l, r) \,|\, l_p \neq 0 \right\}$.

We construct the set $P_p$ of all positively annotated relations and the set $T_{p,k}$ of relations that were predicted with a score from the top $k$ predictions.
Note that we set the fraction in equation \ref{eq:pdd} to $1$ if the denominator is $0$.


\begin{align}
    T_{p,k} &= \left\{(l,r) \in R_p \,|\, r_p<k\right\} \subset R_p
    \\
    P_p &= \left\{(l,r) \in R_p \,|\, l_p = 1 \right\} \subset R_p
    \\
    PDO_p &\mathrel{:}= 1 - \frac{1}{n-1} \sum_{k=1}^{n-1} \frac{|T_{p,k}\cap P_p|}{|T_{p,k}|}
    \label{eq:pdd}
    \\
    PDD_p &\mathrel{:}= 1 - \frac{1}{n-1} \sum_{k=1}^{n-1} \frac{|T_{p,k}\cap P_p|}{|P_p|}
    \label{eq:pdo}
\end{align}

\begin{table*}[h!]
\begin{center}
    \resizebox{\linewidth}{!}{
    \begin{tabular}{|l|cccc|cccc|cccc|}
    \hline
        & \multicolumn{4}{|c|}{VCTree} & \multicolumn{4}{|c|}{GPSNet} & \multicolumn{4}{|c|}{MOTIFS} \\
        Predicate & P-AUC $\uparrow$ & PDD $\downarrow$ & PDO $\downarrow$ & R@50 $\uparrow$ & P-AUC $\uparrow$ & PDD $\downarrow$ & PDO $\downarrow$ & R@50 $\uparrow$ & P-AUC $\uparrow$ & PDD $\downarrow$ & PDO $\downarrow$ & R@50 $\uparrow$ \\
        \hline\hline
on back of & 0.48 & 0.27 & 0.62 & 0.00 & 0.30 & 0.37 & 0.53 & 0.03 & 0.64 & 0.27 & 0.58 & 0.00 \\
going down & 0.89 & 0.24 & 0.59 & 0.00 & 0.61 & 0.38 & 0.65 & 0.07 & 0.38 & 0.57 & 0.66 & 0.07 \\
painted on & 0.66 & 0.19 & 0.48 & 0.02 & 0.61 & 0.21 & 0.52 & 0.00 & 0.60 & 0.24 & 0.51 & 0.00 \\
lying on & 0.68 & 0.67 & 0.35 & 0.22 & 0.59 & 0.40 & 0.54 & 0.19 & 0.34 & 0.45 & 0.54 & 0.26 \\
jumping over & 0.51 & 0.53 & 0.73 & 0.00 & 0.51 & 0.50 & 0.71 & 0.00 & 0.61 & 0.61 & 0.66 & 0.00 \\
guiding & 0.53 & 0.44 & 0.69 & 0.00 & 0.61 & 0.47 & 0.69 & 0.00 & 0.54 & 0.72 & 0.62 & 0.00 \\
eating & 0.90 & 0.19 & 0.33 & 0.02 & 0.64 & 0.24 & 0.56 & 0.04 & 0.56 & 0.23 & 0.52 & 0.09 \\
drinking & 0.49 & 0.50 & 0.77 & 0.00 & 0.49 & 0.52 & 0.78 & 0.00 & 0.59 & 0.61 & 0.76 & 0.00 \\
catching & 0.46 & 0.74 & 0.48 & 0.00 & 0.39 & 0.65 & 0.64 & 0.00 & 0.35 & 0.87 & 0.60 & 0.00 \\
playing with & 0.91 & 0.44 & 0.51 & 0.00 & 0.80 & 0.39 & 0.73 & 0.00 & 0.83 & 0.60 & 0.42 & 0.00 \\
chasing & 0.42 & 0.72 & 0.71 & 0.00 & 0.40 & 0.53 & 0.60 & 0.00 & 0.23 & 0.63 & 0.63 & 0.00 \\
climbing & 0.67 & 0.84 & 0.58 & 0.00 & 0.72 & 0.94 & 0.35 & 0.00 & 0.68 & 0.68 & 0.65 & 0.00 \\
cleaning & 0.66 & 0.77 & 0.61 & 0.00 & 0.67 & 0.84 & 0.39 & 0.00 & 0.64 & 0.91 & 0.67 & 0.00 \\
pushing & 0.51 & 0.65 & 0.64 & 0.00 & 0.57 & 0.61 & 0.49 & 0.00 & 0.64 & 0.70 & 0.49 & 0.00 \\
pulling & 0.49 & 0.33 & 0.44 & 0.05 & 0.43 & 0.27 & 0.51 & 0.07 & 0.43 & 0.63 & 0.50 & 0.05 \\
opening & 0.63 & 0.90 & 0.51 & 0.00 & 0.43 & 0.92 & 0.40 & 0.00 & 0.60 & 0.84 & 0.54 & 0.00 \\
cooking & 0.62 & 0.83 & 0.48 & 0.00 & 0.59 & 0.77 & 0.39 & 0.00 & 0.63 & 0.84 & 0.54 & 0.00 \\
throwing & 0.51 & 0.62 & 0.72 & 0.00 & 0.42 & 0.58 & 0.63 & 0.00 & 0.57 & 0.58 & 0.63 & 0.08 \\
slicing & 0.50 & 0.45 & 0.85 & 0.00 & 0.50 & 0.43 & 0.83 & 0.00 & 0.37 & 0.72 & 0.76 & 0.00 \\
about to hit & 0.70 & 0.26 & 0.49 & 0.67 & 0.73 & 0.33 & 0.38 & 0.70 & 0.64 & 0.35 & 0.56 & 0.74 \\
kicking & 0.67 & 0.85 & 0.45 & 0.25 & 0.59 & 0.69 & 0.56 & 0.00 & 0.65 & 0.84 & 0.62 & 0.25 \\
swinging & 0.83 & 0.46 & 0.23 & 0.08 & 0.73 & 0.62 & 0.35 & 0.16 & 0.76 & 0.73 & 0.46 & 0.16 \\
entering & 0.57 & 0.45 & 0.55 & 0.00 & 0.61 & 0.45 & 0.47 & 0.00 & 0.52 & 0.43 & 0.55 & 0.00 \\
exiting & 0.49 & 0.70 & 0.63 & 0.00 & 0.49 & 0.73 & 0.55 & 0.00 & 0.58 & 0.42 & 0.73 & 0.00 \\
enclosing & 0.99 & 0.05 & 0.70 & 0.03 & 0.90 & 0.12 & 0.62 & 0.03 & 0.72 & 0.11 & 0.72 & 0.07 \\
leaning on & 0.62 & 0.15 & 0.74 & 0.00 & 0.78 & 0.13 & 0.71 & 0.00 & 0.68 & 0.18 & 0.77 & 0.00 \\
\hline
\textbf{mean} & 0.63 & 0.51 & 0.57 & 0.05 & 0.58 & 0.50 & 0.56 & 0.05 & 0.57 & 0.57 & 0.60 & 0.07 \\
\hline
    \end{tabular}
    }
    \end{center}
    \caption{Metric results for three different scene graph generation models, evaluated on the Haystack dataset for the predicate classification task. We add the Recall@50 metric, calculated on the PSG test set for reference. For Predicate Discrimination Disadvantage (PDD) and Predicate Dominance Overestimation (PDO), lower scores are better. For Predicate ROC-AUC (P-AUC) and Recall@k (R@50), higher scores are better. The bottom row is the average over all rows above and represents a unified score for the whole dataset.}
    \label{tab:scores}
\end{table*}

A high PDD score results when the predicate appears rarely in the top scores but would have been expected to be there, \ie when it is displaced by other predicates.
A high PDO score appears when a predicate is too often in the top scores but is not expected there, \ie it displaces other predicates.
Note that PDD and PDO go hand in hand and should always be evaluated together.

PDD and PDO are defined using recall and precision scores respectively and are therefore robust against unbalanced labels. A metric susceptible to the positive-negative ratio would return skewed results because the Haystack dataset contains varying amounts of negative ratios depending on the predicate class.

\section{Experiments}

We evaluate the top 3 performing predicate classification models \cite{gpsnet,vctree,motifs} (the ResNet-101 variant) from the PSG paper \cite{psg} and report our proposed metrics on the Haystack dataset. We cannot evaluate our metrics on the PSG test set because then we would rely on implicit negative annotations that inevitably perturb the metrics.

In table \ref{tab:scores}, we report metric scores for selected rare predicate classes that are present in the Haystack dataset. Additionally, we show the R@50 score on the PSG test set for reference. Finally, we report the mean value over all predicate classes for each metric.
For most rare predicate classes, R@50 returns the lowest possible value of 0.0, making it virtually impossible to derive any interesting information from it. In contrast, our methods report different values even for predicate classes that are difficult to predict.

The correlation between P-AUC and R@50 is about 0.22, indicating that the two metrics are indeed looking at different aspects of the model output. Some predicates like "eating" or "playing with" have very low scores on R@50 but very high on P-AUC. This indicates that such predicates are understood by the model but rarely make it to the top 50 predicates. For "playing with", this could for example happen where many people are playing together or interactions between other objects in the image are deemed more important.

Values for PDO are expected to be low for rare predicate classes which usually don't displace other predicates. The highest ranked predicate with PDO is "slicing", which makes sense because subjects and objects that are in a "slicing" relation with each other usually don't have many alternative predicates that would make sense. This information could not be derived from the R@50 metric.

In general, VCTree performs best when evaluated with a standard mR@50 compared to the other two methods and performs best on our three metrics as well. It has the highest Predicate ROC-AUC score, indicating the best understanding of rare predicate classes and the lowest PDD and PDO scores, which demonstrate that the model is more suitable at deciding between predicates within a relation. The gap to GPSNet is small though and GPSNet could be improved by focusing more on the predicates independently.

With the P-AUC, we can see that existing models are indeed capable of understanding rare predicate classes like "playing with". The R@50 metric does not provide this kind of information and only tells us that the predicates loose against other predicates on the image. However, with the PDD and PDO metrics, we can detect that this problem already occurs at a relation level. Future scene graph model architectures should take this into account and improve their predicate ranking on individual relations. Existing models appear to already have a fundamental understanding of the individual predicate classes.

\section{Conclusion}

We presented the Haystack scene graph dataset and showed how our annotation pipeline is specifically designed to assist existing scene graph datasets with rare predicate classes. We use a model-assisted approach to streamline the annotation process and generate as many rare predicates as fast as possible.
The Haystack dataset enables us to develop new scene graph metrics that are tweaked towards deeper relation-level insights into model predictions, with a focus on rare predicate classes. With reliable negative annotations available, many metrics from other fields in computer vision and statistics can be applied to the scene graph context.
In the future, we will continue our research in this direction and increase the size and quality of our dataset. We are excited to see how other authors will integrate explicit negatives from our dataset into their work.



{\small
\bibliographystyle{ieee_fullname}
\bibliography{egbib}

\begin{thebibliography}{10}\itemsep=-1pt

\bibitem{survey2023}
Xiaojun Chang, Pengzhen Ren, Pengfei Xu, Zhihui Li, Xiaojiang Chen, and Alex
  Hauptmann.
\newblock A comprehensive survey of scene graphs: Generation and application.
\newblock {\em {IEEE} Transactions on Pattern Analysis and Machine
  Intelligence}, 45(1):1--26, jan 2023.

\bibitem{devil_in_tails}
Alakh Desai, Tz-Ying Wu, Subarna Tripathi, and Nuno Vasconcelos.
\newblock Learning of visual relations: The devil is in the tails.
\newblock pages 15404--15413.

\bibitem{dong_stacked}
Xingning Dong, Tian Gan, Xuemeng Song, Jianlong Wu, Yuan Cheng, and Liqiang
  Nie.
\newblock Stacked hybrid-attention and group collaborative learning for
  unbiased scene graph generation.
\newblock pages 19427--19436, 2022.

\bibitem{sam}
Alexander Kirillov, Eric Mintun, Nikhila Ravi, Hanzi Mao, Chloe Rolland, Laura
  Gustafson, Tete Xiao, Spencer Whitehead, Alexander~C. Berg, Wan-Yen Lo, Piotr
  Dollár, and Ross Girshick.
\newblock Segment {Anything}, Apr. 2023.
\newblock arXiv:2304.02643 [cs].

\bibitem{visualgenome}
Ranjay Krishna, Yuke Zhu, Oliver Groth, Justin Johnson, Kenji Hata, Joshua
  Kravitz, Stephanie Chen, Yannis Kalantidis, Li-Jia Li, David~A. Shamma,
  Michael~S. Bernstein, and Li Fei-Fei.
\newblock Visual genome: Connecting language and vision using crowdsourced
  dense image annotations.
\newblock 123(1):32--73, 2017.

\bibitem{maskdino}
Feng Li, Hao Zhang, Huaizhe Xu, Shilong Liu, Lei Zhang, Lionel~M Ni, and
  Heung-Yeung Shum.
\newblock Mask {DINO}: {Towards} a {Unified} {Transformer}-{Based} {Framework}
  for {Object} {Detection} and {Segmentation}.

\bibitem{gpsnet}
Xin Lin, Changxing Ding, Jinquan Zeng, and Dacheng Tao.
\newblock Gps-net: Graph property sensing network for scene graph generation.
\newblock {\em 2020 IEEE/CVF Conference on Computer Vision and Pattern
  Recognition (CVPR)}, pages 3743--3752, 2020.

\bibitem{lu2016visual}
Cewu Lu, Ranjay Krishna, Michael Bernstein, and Li Fei-Fei.
\newblock Visual relationship detection with language priors.
\newblock In {\em European Conference on Computer Vision}, 2016.

\bibitem{pixels2graph}
Alejandro Newell and Jia Deng.
\newblock Pixels to graphs by associative embedding.
\newblock {\em Advances in Neural Information Processing Systems},
  2017-December:2172--2181, 2017.
\newblock 31st Annual Conference on Neural Information Processing Systems, NIPS
  2017 ; Conference date: 04-12-2017 Through 09-12-2017.

\bibitem{dinov2}
Maxime Oquab, Timothée Darcet, Théo Moutakanni, Huy Vo, Marc Szafraniec,
  Vasil Khalidov, Pierre Fernandez, Daniel Haziza, Francisco Massa, Alaaeldin
  El-Nouby, Mahmoud Assran, Nicolas Ballas, Wojciech Galuba, Russell Howes,
  Po-Yao Huang, Shang-Wen Li, Ishan Misra, Michael Rabbat, Vasu Sharma, Gabriel
  Synnaeve, Hu Xu, Hervé Jegou, Julien Mairal, Patrick Labatut, Armand Joulin,
  and Piotr Bojanowski.
\newblock {DINOv}2: Learning robust visual features without supervision.

\bibitem{vctree}
Kaihua Tang, Hanwang Zhang, Baoyuan Wu, Wenhan Luo, and Wei Liu.
\newblock Learning to compose dynamic tree structures for visual contexts.
\newblock In {\em 2019 {IEEE}/{CVF} Conference on Computer Vision and Pattern
  Recognition ({CVPR})}, pages 6612--6621.
\newblock {ISSN}: 2575-7075.

\bibitem{imp}
Danfei Xu, Yuke Zhu, Christopher~B. Choy, and Li Fei-Fei.
\newblock Scene graph generation by iterative message passing.

\bibitem{pcpl}
Shaotian Yan, Chen Shen, Zhongming Jin, Jianqiang Huang, Rongxin Jiang, Yaowu
  Chen, and Xian-Sheng Hua.
\newblock {PCPL}: Predicate-correlation perception learning for unbiased scene
  graph generation.
\newblock In {\em Proceedings of the 28th {ACM} International Conference on
  Multimedia}, {MM} '20, pages 265--273. Association for Computing Machinery.

\bibitem{psg}
Jingkang Yang, Yi~Zhe Ang, Zujin Guo, Kaiyang Zhou, Wayne Zhang, and Ziwei Liu.
\newblock Panoptic scene graph generation.

\bibitem{motifs}
Rowan Zellers, Mark Yatskar, Sam Thomson, and Yejin Choi.
\newblock Neural motifs: Scene graph parsing with global context.
\newblock In {\em Conference on Computer Vision and Pattern Recognition}, 2018.

\bibitem{IETrans}
Ao Zhang, Yuan Yao, Qianyu Chen, Wei Ji, Zhiyuan Liu, Maosong Sun, and Tat-Seng
  Chua.
\newblock Fine-grained scene graph generation with data transfer.

\bibitem{HiLo}
Zijian Zhou, Miaojing Shi, and Holger Caesar.
\newblock {HiLo}: Exploiting high low frequency relations for unbiased panoptic
  scene graph generation.

\bibitem{survey2022}
Guangming Zhu, Liang Zhang, Youliang Jiang, Yixuan Dang, Haoran Hou, Peiyi
  Shen, Mingtao Feng, Xia Zhao, Qiguang Miao, Syed Afaq~Ali Shah, and
  Bennamoun.
\newblock Scene graph generation: A comprehensive survey.
\newblock {\em ArXiv}, abs/2201.00443, 2022.

\end{thebibliography}
}

\end{document}